\documentclass{article}
\usepackage[preprint]{tmlr}   

\usepackage{amsmath,amssymb}
\usepackage{booktabs}
\usepackage{graphicx}
\usepackage{xcolor}
\usepackage{pgfplots}
\pgfplotsset{compat=1.16}
\usepackage{hyperref}
\hypersetup{hidelinks}
\usepackage{url}
\definecolor{oiBlue}{HTML}{0072B2}
\definecolor{oiOrange}{HTML}{E69F00}
\definecolor{oiVermillion}{HTML}{D55E00}
\definecolor{oiGreen}{HTML}{009E73}

\title{Calibrated e-CUSUM Decoding for Quantized Reasoning Models:\\
Why Token Log-Probability Is the Wrong Observable for Decoding Monitors}

\author{%
  \name El Hassane Ettifouri \email eettifouri@novelis.io \\
  \addr Novelis Research, Paris, France \\
  ORCID: \url{https://orcid.org/0000-0001-5299-9053}
  \AND
  \name Ayoub Belfatmi \email abelfatmi@novelis.io \\
  \addr Novelis Research, Paris, France \\
  ORCID: \url{https://orcid.org/0009-0005-4010-794X}
  \AND
  \name Mahaman Sanoussi Yahaya Alassan \email syahaya@novelis.io \\
  \addr Novelis Research, Paris, France \\
  ORCID: \url{https://orcid.org/0009-0006-0825-4701}
  \AND
  \name Walid Dahhane \email wdahhane@novelis.io \\
  \addr Novelis Research, Paris, France \\
  ORCID: \url{https://orcid.org/0000-0001-5387-3380}
}

\begin{document}
\maketitle

\begin{abstract}
Low-bit quantization makes small reasoning models cheap to deploy but degrades
their chain-of-thought, and a recurring folk remedy is to \emph{monitor} the
decoder and intervene when generation ``goes wrong''. A natural candidate for
such a monitor is a martingale built from token log-probabilities, controlled by
a concentration inequality. We show, both analytically and empirically, that this
observable is the wrong basis for a decoder monitor: the increment $\log p(w_t)+H_t$ is a
mean-zero martingale \emph{by construction under the model's own sampling law},
so it is blind to exactly the failure it is meant to catch---confident
repetition---and, when naively checked throughout generation, it does not provide
a useful sequential alarm. We replace it with a training-free decoding controller
built on two ingredients that are individually standard but, to our knowledge,
not previously combined for this purpose: (i) a \emph{degeneration-aware} alarm
score that fuses token uncertainty with an explicit verbatim-repetition signal,
and (ii) a calibrated e-process-based sequential detector. The raw product
process is Ville-valid under a conditional-mean null; the deployed CUSUM-floored
statistic is treated more cautiously as an empirical change detector because the
alarm score is history-dependent and autocorrelated. On GSM8K with
DeepSeek-R1-Distill-Qwen-1.5B, the controller reduces the observed
verbatim-degeneration signals in this pilot and, once calibrated, becomes a
selective detector of failing traces
($\phi\!\approx\!0.3$, precision $\approx\!0.6$ versus a $0.38$ base rate) rather
than firing on $93\%$ of generations as the uncalibrated version does. We report a
positive but statistically \emph{inconclusive} accuracy trend (INT4:
$63\%\!\rightarrow\!69\%$, paired McNemar $p=0.18$ at $n{=}100$), an honest
token-budget cost of $+28\%$, and the observation---counter to the motivating
intuition---that on GSM8K the dominant failure of these models is
non-termination, not looping. We release the decoder, the calibration and
re-scoring tools, and all traces. This is a preliminary study whose main
contribution is methodological: a clear account of why a tempting token-level
observable is inadequate, and a calibrated, honestly-evaluated replacement.
\end{abstract}

\section{Introduction}

Distilled reasoning models such as DeepSeek-R1-Distill \citep{deepseekr1} put
competitive mathematical reasoning within reach of commodity hardware, and
4-bit post-training quantization pushes them further, onto a single consumer GPU
or even a CPU. The price is well documented: low-bit quantization degrades
mathematical reasoning disproportionately, inflating method and execution errors
and lengthening chains of thought, with errors that tend to appear early and
cascade \citep{quantreasoning,quanthurts}. This points to a different intervention point. Rather than waiting for a finished answer and re-ranking it, one can act while the reasoning trace is still being generated. Because a handful of early tokens is often enough to send a quantized model down a bad path, a decoder-side monitor may catch the trajectory and redirect it before the mistake propagates into the final answer.

The obvious way to formalize this is as sequential testing. The decoder hands
us a full distribution at each step, so it is tempting to form a token-level
statistic, sum it as generation proceeds, and raise an alarm once a martingale
deviation bound is crossed. Concretely, one takes the token log-probability
$X_t = \log p(w_t \mid w_{<t})$ as the running quantity, subtracts its
conditional expectation, and flags the trajectory when the cumulative sum
drifts outside an Azuma--Hoeffding band \citep{hoeffding1963}. 
This is appealing, it sounds rigorous, and---as we show---it
does not work as a decoding monitor. More precisely, the martingale formalism is
not the problem: the problem is the \emph{observable}. Centering a token
log-probability under the model's own sampling law produces a valid martingale,
but one that measures self-consistency of sampling rather than trajectory health.
The purpose of this paper is first to explain \emph{why} this observable cannot
work, precisely, and then to give a replacement that is both defensible and
honestly evaluated.

Our analysis is elementary but, we think, clarifying. When $w_t$ is sampled from
the same (tempered, truncated) distribution used to compute the Shannon entropy
$H_t$, the increment $D_t=X_t-\mathbb{E}[X_t\mid\mathcal{F}_{t-1}]=\log p(w_t)+H_t$
has conditional mean exactly zero. The cumulative sum $M_n=\sum_t D_t$ is
therefore a genuine martingale---but a martingale \emph{with respect to the
model's own generative law}, which measures the self-consistency of sampling and
carries no information about correctness. In a confident repetition loop the
chosen token has $p\!\approx\!1$, so $\log p\!\approx\!0$ and $H_t\!\approx\!0$,
giving $D_t\!\approx\!0$: the monitor is silent precisely when the model is
confidently wrong. Azuma's band, in turn, reacts only to \emph{high-variance}
(high-entropy) stretches, i.e.\ to uncertainty, not to error.

We take this failure seriously as a design constraint and build a controller that
avoids it (Section~\ref{sec:method}). It rests on two pieces. First, an alarm
score $a_t\in[0,1]$ that combines normalized token uncertainty with an explicit
\emph{verbatim-repetition} signal; the repetition channel is what lets the monitor
see confident loops that the log-probability martingale cannot. Second, a
calibrated e-process-based detector. The raw product
$E_n^{\mathrm{raw}}=\prod_t\bigl(1+\lambda(a_t-\mu_0)\bigr)$ has a genuine
Ville-style, time-uniform false-alarm guarantee under the conditional null
$\mathbb{E}[a_t\mid\mathcal{F}_{t-1}]\le\mu_0$. The implemented controller uses a
CUSUM-floored log statistic for change detection; $a_t$ being strongly
history-dependent, we report its operating point empirically rather than treating
it as an exact e-process. Far from a technicality, this calibration does most of the work: a badly
chosen baseline makes the alarm trip on nearly everything, while a calibrated
one stays quiet except on the traces that genuinely warrant a closer look.

Our evaluation uses GSM8K \citep{gsm8k} with DeepSeek-R1-Distill-Qwen-1.5B, in
both FP16 and INT4. We read the outcomes conservatively: they do not all line
up, but taken together they point to the following:

\begin{itemize}
\item \textbf{Token log-probability is the wrong signal for this monitor.}
It overlooks confident loops in our sanity check, and on real traces the
uncalibrated version alarms on nearly every generation ($93$--$95\%$;
Sections~\ref{sec:why} and~\ref{sec:results}).
\item \textbf{Calibration is what buys selectivity here.} Fixing $\mu_0$ at
the $90^{\text{th}}$ percentile of $a_t$ over healthy FP16 traces converts an
indiscriminate alarm into one reaching $\phi\!\approx\!0.3$ and precision
$\approx\!0.6$, against a $0.38$ base rate.
\item \textbf{The controller reduces the verbatim-degeneration signals we
measure} in this pilot (INT4 mean consecutive repetition falls from $0.010$
to $0.003$; severe loops from $1\%$ to $0\%$) and trims non-termination a
little. With severe loops so infrequent at $n=100$, we read this as
descriptive rather than as a population-level result.
\item \textbf{The accuracy effect is positive but not yet significant}
($68\!\rightarrow\!72$ FP16, $63\!\rightarrow\!69$ INT4; paired McNemar
$p=0.48$ and $p=0.18$), at a token cost of $+28\%$. We do not claim an accuracy
improvement on this evidence.
\item \textbf{The motivating intuition is partly wrong.} On GSM8K, verbatim loops
are rare ($\le 4\%$ of tokens even on failing traces); the dominant failure mode
is non-termination (up to $49\%$ of incorrect INT4 traces exhaust the budget), a
phenomenon independently reported by \citet{lotfi2606}.
\end{itemize}

We view this as a preliminary study. Its lasting contribution is methodological: a
precise account of why an attractive token-level observable is inadequate, and a
calibrated empirical e-CUSUM alternative evaluated without overclaiming. All code, calibration
tools, and traces are released to support the larger-scale study the accuracy
question ultimately needs.

\section{Related Work}

\paragraph{Quantization and reasoning.} Several recent studies quantify how
low-bit post-training quantization erodes reasoning, showing sharp accuracy drops,
longer chains of thought, and early cascading errors under 4-bit settings
\citep{quantreasoning,quanthurts}; the observation that errors appear early and
propagate is what motivates intervening \emph{during} decoding rather than
re-ranking completed generations. Most directly related to our own findings,
\citet{lotfi2606} report that quantized models over-sample ``overthinking''
markers (\emph{wait}, \emph{but}, \emph{alternatively}) at high-entropy positions
and, in up to $52\%$ of failures, reach the correct answer in intermediate steps
yet never commit to it; they mitigate this with a training-free logit penalty on
those markers. Our independent finding that non-termination dominates INT4
failures on GSM8K corroborates theirs, and we return in
Section~\ref{sec:results} to a tension it creates for marker-based remediation.
These works motivate an inference-time remedy that avoids the expensive
quantization-aware retraining they also discuss.

\paragraph{Confidence- and entropy-based test-time control.} A growing line uses
model-internal signals to steer or filter reasoning. DeepConf \citep{deepconf}
aggregates token-level confidence to filter whole traces before majority voting.
The reasoning-path deviation monitor \citep{rpdi} is the closest prior work to
ours: it is likewise training-free and reads token entropy online, but it thresholds
a fixed local-to-global entropy ratio and only \emph{terminates} the reasoning
block, whereas we add a signal entropy cannot supply (verbatim repetition),
\emph{backtrack and re-explore} rather than terminate, and calibrate an empirical
false-alarm operating point. Work on underthinking \citep{underthinking} and on entropy
minimization \citep{entropymin} likewise ties entropy to reasoning quality. In
short, our controller differs from this family on three axes: it acts \emph{within}
a single trace by backtracking rather than filtering or terminating; its alarm
score is \emph{degeneration-aware}, not entropy-only; and detection is a
calibrated sequential test rather than a fixed threshold.

\paragraph{Degeneration and self-correction.} Neural text degeneration and the
self-reinforcing nature of repetition are long-studied \citep{holtzman2020}, with
data-centric \citep{repetitionout} and decoding-time \citep{penaltydecoding}
accounts of the self-reinforcement that drives it, and recent analyses specific to
reasoning models \citep{circular}. This literature is the reason we do not rely on
entropy alone: varied text has \emph{high} entropy, so an entropy monitor cannot
flag a confident loop, which is precisely when a dedicated repetition signal is
needed. On the intervention side, appending correction markers such as ``Wait''
can elicit self-correction without fine-tuning
\citep{selfcorrectionbench,selfbacktracking}, and truncated sampling schemes such
as min-p \citep{minp} preserve fluency better than raising temperature---indeed,
inflating temperature on an anomaly, as an earlier version of our own decoder did,
increases randomness exactly when the model is already off-track. Our remediation uses these building blocks. Self-consistency
\citep{selfconsistency} is the standard compute-scaling baseline we compare
against.

\paragraph{Anytime-valid inference.} Testing by betting and game-theoretic
statistics provide sequential tests that remain valid under optional stopping via
Ville's inequality \citep{ville1939,shafer2021testing,ramdas2023gametheoretic},
and conformal test martingales instantiate this for exchangeability monitoring
\citep{vovk2005algorithmic}. Closest in application, WATCH monitors deployed models
online with weighted-conformal test martingales \citep{watch}, and adaptive
conformal inference by betting controls miscoverage sequentially
\citep{adaptiveconformalbetting}. We adapt this anytime-valid machinery to the token stream of a single decoder,
while keeping a clear distinction between the raw e-process, which is Ville-valid
under its conditional null, and the deployed e-CUSUM statistic, whose threshold is
calibrated and reported empirically. Restart-valid e-detectors, which combine
e-processes started at successive times and can yield nonasymptotic false-alarm
control for sequential change detection \citep{shin2024edetectors}, are the
natural formal route for turning our practical CUSUM-style monitor into a fully
valid change detector.

\section{Method}
\label{sec:method}

\subsection{An alarm score that sees degeneration}

At each step the decoder produces logits, from which we form a sampling
distribution using temperature and min-p truncation \citep{minp}. From this
distribution we read the Shannon entropy $H_t$ and, after sampling $w_t$, its
log-probability. At each step, the controller reduces the decoder state to a bounded alarm score
$a_t\in[0,1]$:
\begin{equation}
a_t=\min\!\bigl(1,\; w_{\text{rep}}\,r_t + w_{\text{ent}}\,u_t\bigr),
\qquad w_{\text{rep}}=0.7,\; w_{\text{ent}}=0.3 .
\end{equation}
The term $u_t$ flags entropy spikes relative to the recent history of $H_t$.
The term $r_t$ flags verbatim repetition by looking, in a sliding window, at how
much of the text is covered by recurring $n$-grams. Crucially, $r_t$ approaches $1$ during a loop even when the
distribution is peaked and $H_t\!\approx\!0$; this is the signal that a
log-probability martingale structurally cannot provide.

\subsection{From a raw e-process to an implemented e-CUSUM detector}

We monitor $a_t$ using a betting construction \citep{shafer2021testing}. Given a
baseline level $\mu_0$ (the ``healthy'' mean of $a_t$; see
Section~\ref{sec:calib}) and a betting fraction $\lambda\in(0,1)$, define the raw
nonnegative product process
\begin{equation}
E_n^{\mathrm{raw}}=\prod_{t=1}^{n}\bigl(1+\lambda\,(a_t-\mu_0)\bigr),
\qquad
\log E_n^{\mathrm{raw}}=
\sum_{t=1}^{n}\log\bigl(1+\lambda(a_t-\mu_0)\bigr).
\end{equation}
Because $a_t\in[0,1]$, $\mu_0\in[0,1]$, and $\lambda\in(0,1)$, each factor is
positive. Under the conditional null that generation is healthy in the sense
$\mathbb{E}[a_t\mid\mathcal{F}_{t-1}]\le\mu_0$ at every token, each factor has
conditional mean at most $1$. Thus $E_n^{\mathrm{raw}}$ is a nonnegative
supermartingale and Ville's inequality gives the time-uniform guarantee
\begin{equation}
\Pr\!\bigl(\exists n:\; E_n^{\mathrm{raw}}\ge 1/\delta\bigr)\le\delta .
\label{eq:ville}
\end{equation}
Unlike an Azuma band tested every $k$ tokens, \eqref{eq:ville} is valid when the
raw process is queried at \emph{every} token.

The deployed detector uses a CUSUM-floored log statistic,
\begin{equation}
S_0=0,\qquad
S_n=\max\!\left(0,\; S_{n-1}+\log\bigl(1+\lambda(a_n-\mu_0)\bigr)\right),
\label{eq:cusum}
\end{equation}
and raises an alarm when $S_n\ge\tau$. This floor makes the controller target
sustained rises of $a_t$ above the healthy baseline rather than banking negative
drift. We use it as an e-process-inspired change detector, not as a literal
supermartingale: the reset in \eqref{eq:cusum}, together with the history-dependence
of $a_t$, means that its false-alarm behaviour must be calibrated and reported
empirically.
\subsection{Offline calibration of the baseline}
\label{sec:calib}

Guarantee~\eqref{eq:ville} is only meaningful if $\mu_0$ is a true upper bound on
the healthy conditional mean of $a_t$. Setting $\mu_0$ too low makes the product
process drift upward on \emph{all} traces and destroys selectivity---this, we will
see, is exactly what happened in our first run. We therefore calibrate $\mu_0$
offline: we take healthy reference traces (correct and non-truncated FP16
generations), pool their per-token $a_t$, and set $\mu_0$ to a high quantile (we
use the $90^{\text{th}}$ percentile). This is a pragmatic marginal calibration,
not a certificate of the conditional bound in \eqref{eq:ville}. Because $a_t$ is
computed from running entropy statistics and sliding-window repetition, it is
strongly history-dependent and autocorrelated; there can be decoder states for
which $\mathbb{E}[a_t\mid\mathcal{F}_{t-1}]>\mu_0$ even if the marginal healthy
level is below $\mu_0$. We therefore treat the false-alarm rate of the implemented
e-CUSUM detector as an empirical operating point to be reported, not as a direct
consequence of Ville's inequality.
\subsection{Remediation}

When $S_n$ crosses the calibrated threshold $\tau$, the controller backtracks by
truncating the last $k$ tokens and rebuilding a clean key--value cache, then
resumes with a graded, training-free response inspired by the self-correction
literature \citep{selfcorrectionbench,selfbacktracking}: it widens or tightens
min-p rather than inflating temperature \citep{minp}, and on repeated alarms it
injects a short correction marker (``\texttt{Wait, let me re-check the previous
step.}''). A budget caps the number of interventions. Normal steps use the
incremental KV cache; only the (infrequent) backtracks recompute, which keeps the
per-token cost close to standard decoding.

These roles are kept separate in the implementation: a first module reads the
logits and recent tokens and emits $(u_t,r_t,a_t)$; a second maintains both
the raw product and the CUSUM statistic; a third carries out the
intervention itself---truncation, cache rebuild, and the sampling-parameter
changes. The same trace can therefore be replayed for detector evaluation
without rerunning the model or applying the intervention.
\section{Why token log-probability is the wrong observable}
\label{sec:why}

We can see the limitation of $\log p(w_t)+H_t$ directly in a simple controlled
sanity check: it does not react reliably when the model enters a confident loop.
We construct synthetic token streams for four regimes calibrated to real
behaviour: \emph{healthy} (moderate entropy, varied tokens), \emph{confident
loop} (peaked distribution, $\log p\!\approx\!0$, verbatim repetition),
\emph{uncertain drift} (sustained high entropy, no repetition), and
\emph{drift-then-collapse} (rising uncertainty that collapses into a loop---the
trajectory we observe on trick prompts). We run the log-probability martingale
(the Azuma monitor of the earlier decoder) and our calibrated e-CUSUM detector on
each and record the detection rate. This experiment is not intended as a
competitive benchmark: the regimes are constructed to isolate the mechanism, in
particular the difference between entropy-only monitoring and an explicit
repetition channel.

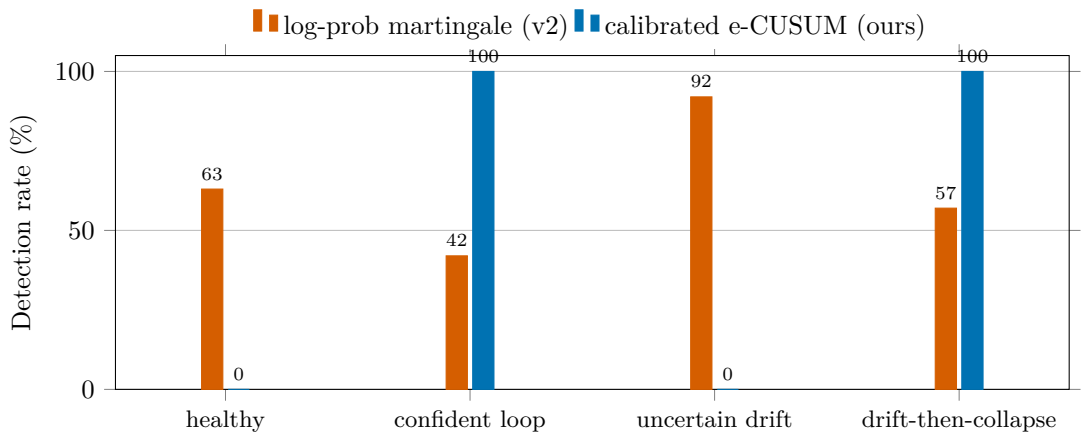
\begin{figure}[t]
\centering
\begin{tikzpicture}
\begin{axis}[
    ybar, bar width=8pt,
    width=0.86\textwidth, height=6cm,
    ymin=0, ymax=105,
    ylabel={Detection rate (\%)},
    symbolic x coords={healthy, confident loop, uncertain drift, drift-then-collapse},
    xtick=data, x tick label style={font=\small},
    enlarge x limits=0.15,
    legend style={at={(0.5,1.02)}, anchor=south, legend columns=2, draw=none},
    nodes near coords, nodes near coords style={font=\scriptsize},
    ymajorgrids, tick align=outside,
]
\addplot[fill=oiVermillion, draw=oiVermillion] coordinates
  {(healthy,63) (confident loop,42) (uncertain drift,92) (drift-then-collapse,57)};
\addplot[fill=oiBlue, draw=oiBlue] coordinates
  {(healthy,0) (confident loop,100) (uncertain drift,0) (drift-then-collapse,100)};
\legend{log-prob martingale (v2), calibrated e-CUSUM (ours)}
\end{axis}
\end{tikzpicture}
\caption{Controlled sanity check on synthetic token streams ($100$ per regime). The
log-probability martingale both cries wolf on healthy text and misses confident
loops; the calibrated e-CUSUM abstains on healthy and pure-uncertainty streams and
catches the two constructed degeneration regimes. For \emph{healthy} and \emph{uncertain drift} a low
bar is the desired behaviour.}
\label{fig:synthetic}
\end{figure}

Figure~\ref{fig:synthetic} shows the outcome of this sanity check. The log-probability martingale
raises an alarm on $63\%$ of \emph{healthy} streams---a false-alarm rate that
already disqualifies it---yet detects the \emph{confident loop} only $42\%$ of the
time, and that only by the accident of a random walk crossing its band. The
calibrated detector, by contrast, does not fire on healthy or on pure uncertain
drift, and detects the confident loop and the realistic collapse in $100\%$ of
these constructed cases. This is not evidence of broad empirical dominance; it is
a mechanism check. The asymmetry is not a tuning artifact: it follows from the
analysis above. A monitor
centered on the model's own entropy cannot, in principle, react to a token the
model is confident about, whether that token is right or catastrophically wrong.

\section{Experimental setup}

\paragraph{Model and data.} We use DeepSeek-R1-Distill-Qwen-1.5B \citep{deepseekr1}
in FP16 and in 4-bit NF4 quantization \citep{qlora}, on the first $100$ GSM8K test
problems \citep{gsm8k}, with a $2048$-token generation budget and fixed seeds. We
compare \emph{vanilla} min-p sampling against the same decoder with the controller
enabled (\emph{v3}); the two share sampling code and differ only in whether the
controller may intervene, which isolates its effect. Hardware is a single
NVIDIA RTX A4000.

\paragraph{Controller configuration.} Unless otherwise stated, the alarm score
uses $w_{\mathrm{rep}}=0.7$ and $w_{\mathrm{ent}}=0.3$. The baseline is calibrated
to $\mu_0=0.41$ from healthy FP16 reference traces, after an initial uncalibrated
run used $\mu_0=0.15$. The alarm threshold is parameterized by the nominal
$\delta=0.05$ but interpreted through the empirical false-alarm rate of the CUSUM
statistic. Backtracking, intervention budget, min-p adjustments, and marker
injection are fixed before evaluation and shared across FP16 and INT4 unless
explicitly ablated; the exact configuration files are released with the traces.

\paragraph{Metrics.} We report accuracy (final answer via \verb|\boxed{}| then last
number), cost (mean generated tokens) , and two degeneration measures. 
Mathematical chains of thought being naturally repetitive, a global $n$-gram
coverage statistic is a poor degeneration signal---it exceeds $0.6$ even on correct
answers---so we use instead the \emph{consecutive-repetition rate} (the fraction of
the sequence occupied by the longest verbatim block repeated back-to-back), which
isolates true loops, together with the \emph{truncation rate} (generations that
reach the budget without emitting EOS). We assess accuracy differences with the
paired McNemar test, appropriate for two decoders on the same items.

\section{Results}
\label{sec:results}

\subsection{Main comparison}

Table~\ref{tab:main} collects the run.  The accuracy numbers move in the right direction---$+4$ points in
FP16 and $+6$ in INT4, with INT4+v3 essentially matching FP16 vanilla---but the
paired test does not support a claim of improvement: of the discordant items, the
controller fixes $10$ and breaks $4$ in INT4 (McNemar $p=0.18$) and fixes $11$ and
breaks $7$ in FP16 ($p=0.48$). At $n=100$ this is suggestive at most. The controller also adds $28\%$ to
the token count and $30\%$ to wall-clock time, so any eventual accuracy claim
will have to be made against a compute-matched baseline.

\begin{table}[t]
\centering
\caption{GSM8K, DeepSeek-R1-Distill-Qwen-1.5B, $n=100$, $2048$-token budget.
``Consec.\ rep.'' is the verbatim consecutive-repetition rate; ``Sev.\ loop'' is
the fraction with consec.\ rep.\ $>0.3$; ``Trunc.'' is the no-EOS rate. Accuracy
differences are \emph{not} statistically significant (McNemar $p=0.48$ FP16,
$p=0.18$ INT4).}
\label{tab:main}
\resizebox{\textwidth}{!}{%
\begin{tabular}{llcccccc}
\toprule
Dtype & Method & Acc.\ (\%) & Tokens & Trunc.\ (\%) & Consec.\ rep. & Sev.\ loop (\%) & s/ex \\
\midrule
FP16 & vanilla     & 68.0 & 1174 & 28 & 0.002 & 0 & 15.5 \\
FP16 & v3 (ours)   & 72.0 & 1510 & 31 & 0.003 & 0 & 20.4 \\
INT4 & vanilla     & 63.0 &  997 & 22 & 0.010 & 1 & 13.8 \\
INT4 & v3 (ours)   & 69.0 & 1282 & 19 & 0.003 & 0 & 17.9 \\
\bottomrule
\end{tabular}%
}
\end{table}

\subsection{Observed degeneration and failure modes}

On INT4 the degeneration metrics shift the way we intended, though the
magnitudes stay small. Consecutive repetition drops from $0.010$ to $0.003$,
severe loops from $1\%$ to $0\%$, and truncation from $22\%$ to $19\%$. Since
the severe-loop figure rests on a single trace at $n=100$, we report it as
descriptive and leave its stability to bootstrap intervals or a larger run.
The direction nonetheless matches what the controller is designed to do, and
even its small size is telling: verbatim loops turn out to be rare on GSM8K,
against the intuition that initially motivated the method.
For FP16, consecutive repetition is already at the noise floor ($0.002$ under
vanilla and $0.003$ under v3), so the slight uptick is not a real
degradation; if anything it underlines that these metrics need bootstrap
intervals or larger runs before they can carry population-level claims.
Splitting the vanilla INT4 traces by outcome exposes the true failure mode:
consecutive repetition is $0.002$ on correct answers but $0.024$ on incorrect
ones, and truncation $6\%$ versus $49\%$, with the incorrect answers
consuming far more tokens ($1286$ vs $828$). What breaks this quantized model
on grade-school math is thus not a tight verbatim loop but an inability to
stop---it keeps rambling without ever committing to an answer---consistent
with \citet{lotfi2606}, who find that quantized models often reach the
correct answer in intermediate steps yet fail to emit it, and with the
broader observation that quantization lengthens chains of thought
\citep{quantreasoning}.

\subsection{Calibration is what makes the monitor usable}

\begin{figure}[t]
\centering
\begin{tikzpicture}
\begin{axis}[
    ybar, bar width=13pt,
    width=0.7\textwidth, height=5.6cm,
    ymin=0, ymax=1.05,
    ylabel={Rate (INT4 vanilla traces)},
    symbolic x coords={uncalibrated ($\mu_0{=}0.15$), calibrated ($\mu_0{=}0.41$)},
    xtick=data, x tick label style={font=\small},
    enlarge x limits=0.5,
    legend style={at={(0.5,1.02)}, anchor=south, legend columns=2, draw=none},
    nodes near coords, nodes near coords style={font=\scriptsize},
    ymajorgrids, tick align=outside,
]
\addplot[fill=oiVermillion, draw=oiVermillion] coordinates
  {(uncalibrated ($\mu_0{=}0.15$),0.95) (calibrated ($\mu_0{=}0.41$),0.20)};
\addplot[fill=oiBlue, draw=oiBlue] coordinates
  {(uncalibrated ($\mu_0{=}0.15$),0.93) (calibrated ($\mu_0{=}0.41$),0.46)};
\legend{false-alarm rate (lower better), recall on bad traces (higher better)}
\end{axis}
\end{tikzpicture}
\caption{Offline calibration of $\mu_0$ turns a non-selective alarm (fires on
$\sim\!94\%$ of everything) into a detector that trades a lower empirical
false-alarm rate for meaningful recall. Even after calibration the false-alarm rate stays above the nominal
$\delta=0.05$, which is why we report the CUSUM detector at its empirical
operating point.}
\label{fig:calib}
\end{figure}
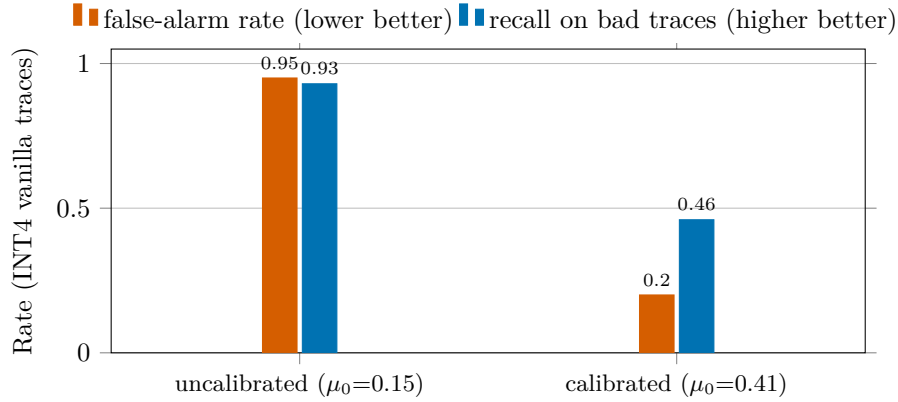

Our first run used an uncalibrated baseline $\mu_0=0.15$. The mean of $a_t$ on the healthy FP16 traces is $0.23$ and its $90^{\text{th}}$
percentile $0.41$, so the baseline was far below the healthy level and the score
accumulation drifted upward on essentially every trace. The consequence, visible
in Figure~\ref{fig:calib}, is a monitor that fires on $93$--$95\%$ of generations
with precision at the base rate---useless as a detector, and actively harmful as a
controller, since indiscriminate intervention is what breaks the $4$--$7$
previously-correct answers noted above. Re-scoring vanilla traces with the
calibrated $\mu_0=0.41$ and the nominal $\delta=0.05$ yields a more selective
detector: on INT4, precision $0.61$ and recall $0.46$ ($\phi=0.28$); on FP16,
precision $0.57$ and recall $0.56$ ($\phi=0.32$), against a $0.38$ base rate of
bad traces. We are candid that the empirical false-alarm rate under calibration
($\approx\!0.20$) still exceeds the nominal $\delta$: the marginal-quantile
calibration does not certify the conditional null of \eqref{eq:ville}, and the
CUSUM floor in \eqref{eq:cusum} is not itself a raw e-process. We therefore
present the method as a calibrated e-CUSUM change detector with a \emph{measured}
operating point rather than one enjoying an exact guarantee at the stated
$\delta$.

\section{Limitations}

This is a single-model, single-dataset, single-seed study at $n=100$; the accuracy
effect is not statistically significant and we do not claim it. The formal Ville
guarantee applies to the raw product process only under the conditional null
$\mathbb{E}[a_t\mid\mathcal{F}_{t-1}]\le\mu_0$ at every token. Our deployed
statistic should be interpreted more cautiously. First, $\mu_0$ is calibrated from
marginal healthy traces, not from a certified upper bound on the conditional mean.
Since $a_t$ is history-dependent and autocorrelated, there may be decoder states
for which $\mathbb{E}[a_t\mid\mathcal{F}_{t-1}]>\mu_0$, so the raw product is not
certified to be a supermartingale on real traces. Second, the CUSUM floor improves
change-detection behaviour but further separates the implemented monitor from a
literal e-process. A strict disjoint calibration/evaluation split, conditional
calibration, a more conservative betting rule, or restart-valid e-detector
construction \citep{shin2024edetectors} is needed before making an exact
statement at the stated $\delta$.
Because the controller spends $28\%$ more tokens, a fair accuracy comparison
needs compute-matched baselines---best-of-$n$ or self-consistency at the same
budget---which we have implemented but not yet run at scale. Until we do,
the positive accuracy trend cannot be told apart from the effect of simply
generating more tokens.The reduction in degeneration is small in absolute terms, since verbatim
loops are uncommon on GSM8K to begin with; turning the observed shifts into
firm statistical claims will take bootstrap intervals and larger runs.
Whether the controller earns its keep on harder benchmarks---MATH
\citep{lightman2023}, AIME, GPQA, where non-termination and drift are more
frequent---is precisely the question left open. Two ablations matter most. The first is to tease apart entropy-only,
repetition-only, and combined alarms on real traces: the synthetic sanity
check isolates the mechanism but does not, on its own, establish what $r_t$
contributes in deployment. The second is to pit backtracking against
plainer termination or early-exit rules, which would sharpen the contrast
with entropy-threshold methods. A sharper tension sits in our remediation
itself: it inserts a ``Wait'' correction marker, whereas \citet{lotfi2606}
show that quantized models already \emph{over-produce} precisely these
markers at high-entropy positions, and that penalizing them helps. Whether
marker injection aids recovery or compounds non-termination is therefore an open
empirical question our current data cannot settle; a marker-free remediation
(min-p modulation alone, or a marker \emph{penalty} in the spirit of
\citealt{lotfi2606}) is a priority ablation. We release the harness so these gaps
can be closed.

\section{Conclusion}

We set out to build a mathematically-grounded monitor for quantized reasoning
decoders and found, first, that the obvious token-level observable---centered
log-probability under the model's own sampling law---is inadequate for the task,
being blind by construction to confident degeneration and non-selective in
practice. We replaced it with a
degeneration-aware, calibrated e-CUSUM controller whose raw betting construction
has an anytime-valid interpretation under a conditional null, and showed that
calibration is the decisive step separating a useless alarm from a selective one. On GSM8K the controller reduces observed verbatim-degeneration signals and points
to non-termination, not looping, as the real failure mode
of these small quantized models; its effect on accuracy is encouraging but, at this
scale, statistically inconclusive. We present the work in that honest register: a
clarifying negative result about a tempting observable, a calibrated empirical
replacement, and an experimental protocol ready to test the accuracy question
where it is most likely to matter.

\section*{Broader Impact}

The method is an inference-time decoding controller that adds no new generative
capability; its aim is to make small, cheaply-deployable reasoning models more
reliable, which has a modest positive efficiency and accessibility angle
(recovering quality without retraining or larger hardware). The main risk of any
reliability tool is over-trust: our monitor controls degeneration and trajectory
drift, which are correlated with but not equivalent to correctness, and it must not
be presented to end users as a guarantee of correct answers. We have been explicit
that the formal statistical guarantee applies only under the stated conditional
null, and that the deployed CUSUM detector is reported through empirical
false-alarm behaviour rather than as a guarantee of answer correctness. We see no specific dual-use or data-privacy concerns
beyond those common to open reasoning models; all data used are public and contain
no personal information.

\section*{Reproducibility Statement}

All components---the decoder, the offline $\mu_0$ calibration, the raw e-process and e-CUSUM
re-scoring metrics (consecutive-repetition and truncation), the synthetic
diagnostic of Section~\ref{sec:why}, and the full token-level traces underlying
Tables and Figures---are released with fixed seeds and exact configurations. The
main run is reproducible on a single consumer GPU.

\bibliographystyle{tmlr}
\bibliography{references}

\end{document}